\documentclass{article}

\PassOptionsToPackage{numbers, compress}{natbib}

\usepackage[preprint]{neurips_2021}

\usepackage[utf8]{inputenc} 
\usepackage[T1]{fontenc}    
\usepackage{hyperref}       
\usepackage{url}            
\usepackage{booktabs}       
\usepackage{amsfonts}       
\usepackage{nicefrac}       
\usepackage{microtype}      
\usepackage{xcolor}         

\usepackage{amsmath,amssymb}
\usepackage{algorithm, algpseudocode}
\usepackage[binary-units=true]{siunitx}

\DeclareMathOperator{\argmax}{argmax}
\usepackage[utf8]{inputenc}
\usepackage{pgfplots}
\usepackage{pgfplotstable}
\pgfplotsset{compat=newest}

\usetikzlibrary{external}

\usepackage{pgfplots}
\usepgfplotslibrary{fillbetween}
\usetikzlibrary{shapes.misc, calc, fit, shapes.geometric}
\newcommand{\gettikzxy}[3]{%
	\tikz@scan@one@point\pgfutil@firstofone#1\relax
	\edef#2{\the\pgf@x}%
	\edef#3{\the\pgf@y}%
}
\usepackage{multirow}
\usepackage{tabularx,booktabs}
\newcolumntype{Y}{>{\centering\arraybackslash}X}
\usepackage[caption=false]{subfig}

\title{Open Set Recognition using Vision Transformer with an Additional Detection Head}

\author{%
	Feiyang Cai\\
	Vanderbilt University\\
	\texttt{feiyang.cai@vanderbilt.edu} \\
	\And
	Zhenkai Zhang \\
	Clemson University \\
	\texttt{zhenkai@clemson.edu} \\
	\And
	Jie Liu \\
	Harbin Institute of Technology \\
	\texttt{jieliu@hit.edu.cn} \\
	\And
	Xenofon Koutsoukos \\
	Vanderbilt University \\
	\texttt{xenofon.koutsoukos@vanderbilt.edu} \\
}

\begin{document}
	
	\maketitle
	\begin{abstract}
	Deep neural networks have demonstrated prominent capacities for image classification tasks in a 
	closed set setting, where the test data come from the same distribution as the training data.
	However, in a more realistic open set scenario, traditional classifiers with 
	incomplete knowledge cannot tackle test data that are not from the training classes.   
	Open set recognition (OSR) aims to address this problem by both identifying unknown classes and distinguishing known classes simultaneously.  
	In this paper, we propose a novel approach to OSR that is based on the vision transformer (ViT) technique.
	Specifically, our approach employs two separate training stages. First, a ViT model is trained 
	to perform closed set classification.
	Then, an additional detection head is attached to the embedded features extracted by the ViT, trained to force the representations of known data to class-specific clusters compactly.
	Test examples are identified as known or unknown based on their distance to the cluster centers.
	To the best of our knowledge, this is the first time to leverage ViT for the purpose of OSR, and our extensive evaluation against several OSR benchmark datasets reveals that our approach significantly outperforms other baseline methods and obtains new state-of-the-art performance.
	The code and trained models are publicly available at \url{https://github.com/feiyang-cai/osr_vit.git}.
	
\end{abstract}
	\section{Introduction}

The rapid development of deep learning techniques over the past few years 
has led to remarkable success in a wide range of application fields.
Deep learning is built upon an underlying assumption that the training and test data come from the same distribution. 
However, in a realistic application, 
the known classes in the training dataset are not complete, and unknown classes might be included during testing.
Such unseen classes appearing in the test data are beyond the knowledge of the closed set classifier and may be incorrectly recognized as one of the known classes.
Open set recognition (OSR)~\cite{DBLP:journals/pami/ScheirerRSB13} aims to build a trustworthy 
recognition system that is able to not only perform accurate classification on the known classes, but also identify and reject an example not yet encountered.   
 
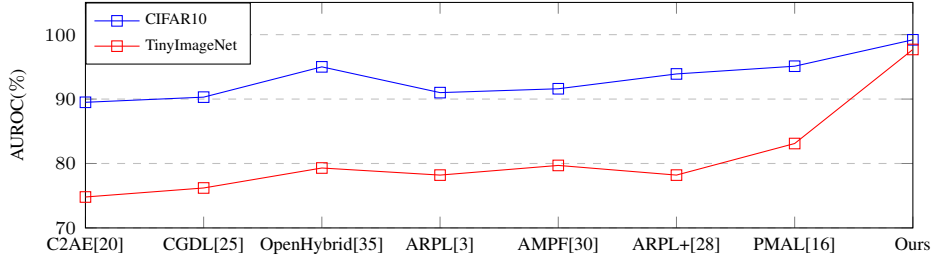
\begin{figure}
	\begin{center}
		    \begin{tikzpicture}
    	\pgfplotsset{compat = newest,		
    		scale only axis,
    		width=11.0cm,
    		height=3.0cm,
    		every tick label/.append style={font=\scriptsize},
    	}
	\begin{axis}[    
		label style={font=\scriptsize},
		tick label style={font=\scriptsize},
		xticklabel style={rotate=0},
		ylabel={AUROC($\%$)},
		xmin=0, xmax=7,
		ymin=70, ymax=105,
		xtick={0,1,2,3,4,5,6,7},
		ytick={ 70, 80, 90,100},
		xticklabels={C2AE\cite{DBLP:conf/cvpr/OzaP19}, CGDL\cite{DBLP:conf/cvpr/SunYZLP20}, OpenHybrid\cite{DBLP:conf/eccv/0002LG020}, ARPL\cite{chen2021adversarial}, AMPF\cite{xia2021adversarial}, ARPL+\cite{vaze2021open}, PMAL\cite{lu2022pmal}, Ours},
		ymajorgrids=true,
		grid style=dashed,
		legend style={font=\tiny, at={(0,1)},anchor=north west},
		legend cell align={left},
		]
		\addplot[
color=blue,
mark=square,
]
coordinates {
	(0,89.5)
	(1,90.3)
	(2, 95.0)
	(3, 91.0)
	(4, 91.6)
	(5, 93.9)
	(6, 95.1)
	(7, 99.2) };

		\addplot[
		color=red,
		mark=square,
		]
		coordinates {
			(0,74.8)
			(1,76.2)
			(2, 79.3)
			(3, 78.2)
			(4, 79.7)
			(5, 78.2)
			(6, 83.1)
			(7, 97.7) };

		\legend{CIFAR10, TinyImageNet}
	\end{axis}
\end{tikzpicture}	
	\end{center}
	\caption{Open set performance trendline on CIFAR10 and TinyImageNet datasets. The methods is in chronological order }
	\label{fig:auroc_trend}
\end{figure}

Since the problem of OSR was initially defined by~\cite{DBLP:journals/pami/ScheirerRSB13},
extensive investigations have been carried out~\cite{DBLP:conf/cvpr/OzaP19,DBLP:conf/cvpr/SunYZLP20,DBLP:conf/eccv/0002LG020,chen2021adversarial,xia2021adversarial,vaze2021open,lu2022pmal}, and over the past years, proposed methods have gradually transitioned from traditional machine learning to deep learning-based methods. 
Although such deep learning-based methods have obtained significant improvement on the standard OSR benchmark datasets, the improvement has slowed down in recent years, as shown in Fig.~\ref{fig:auroc_trend}. 
Typical approaches are based on convolutional neural network (CNN) backbone such as VGG-16~\cite{DBLP:journals/corr/SimonyanZ14a} and ResNet-50~\cite{DBLP:conf/cvpr/HeZRS16}.
Leveraging more recent CNN architectures may be useful to improve the OSR results,
but it might be more helpful to break this performance bottleneck by introducing a new backbone.

Recently, a vision transformer (ViT) model has been proposed in~\cite{DBLP:conf/iclr/DosovitskiyB0WZ21}
using a transformer encoder~\cite{DBLP:conf/nips/VaswaniSPUJGKP17}.  
Based on the self-attention mechanism, which is the core of the transformer encoder, 
the ViT can perform the classification task by making use of global information across the entire image.  
Research results demonstrate that the ViT surpasses the state-of-the-art CNNs and achieves highly competitive performance in benchmarks of several computer vision applications, including image classification~\cite{DBLP:conf/iclr/DosovitskiyB0WZ21,DBLP:conf/icml/TouvronCDMSJ21}, object detection~\cite{DBLP:conf/eccv/CarionMSUKZ20,liu2021Swin}, semantic image segmentation~\cite{jain2021semask,liu2021swinv2}, and action recognition~\cite{DBLP:journals/corr/abs-2201-04288,liu2021swinv2}.
Furthermore, ViT has been used successfully for out-of-distribution detection~\cite{koner2021oodformer} and anomaly detection~\cite{DBLP:conf/isie/MishraVFPF21,DBLP:journals/corr/abs-2111-07677}.
Nevertheless, the strength of ViT has not been explored for solving the OSR problem.

The first and main contribution of the paper is an open set recognition approach based on ViT. 
To the best of our knowledge, this is the first approach using the ViT for OSR.
The proposed approach uses the embedded discriminative features 
(state of the [class] token from the last layer)
extracted from ViT, which can be beneficial not only to classify the known classes but also to identify novel classes.
Specifically, we follow the standard ViT training on the closed set of known classes to retain its dominance in the classification task. 
After the closed set training, 
an additional detection head is attached to the extracted feature space to further force the latent representations to approach class-conditioned cluster centers and reduce intra-class distances, which 
is central to the novelty detection task.
During the testing phase, 
if the distance between the representation of a test sample and its class-specific center exceeds a pre-defined threshold, such a test example will be rejected as an anomaly. Otherwise, the model will recognize it as a known class, which is predicted by the classification head of ViT.     

The paper presents an extensive evaluation of our approach. Specifically, 
the proposed method is evaluated against not only the standard benchmark datasets, but also several more complex datasets, such as large-scale, fine-grained, and long-tailed datasets.  
The evaluation results demonstrate that our method significantly outperforms other existing methods in the literature and achieves new state-of-the-art results on almost all the benchmarks. 
The improvement of open set performance varies from $1.6\%$ in CIFAR+10 to $14.6\%$ in TinyImageNet.


	\section{Related Work}

\subsubsection{Open Set Recognition}
Open set recognition (OSR) has been studied extensively in the literature after it is first formalized in~\cite{DBLP:journals/pami/ScheirerRSB13}.
Most of the early explorations are based on traditional machine learning models, especially support vector machines (SVMs)~\cite{DBLP:journals/pami/ScheirerRSB13,DBLP:conf/eccv/JainSB14}.  
With the development of deep learning techniques, OSR was addressed using deep learning models
such as the OpenMax model introduced in~\cite{DBLP:conf/cvpr/BendaleB16}
replacing the SoftMax layer with an OpenMax layer to enable the model to predict unknown classes. 
Class conditioned autoencoder (C2AE)~\cite{DBLP:conf/cvpr/OzaP19} and conditional Gaussian distribution learning (CGDL)~\cite{DBLP:conf/cvpr/SunYZLP20} are two important reconstructed-based approaches for OSR. 
C2AE employs a class-conditioned autoencoder to reconstruct the input and uses the reconstruction error to identify the unknown class.
CGDL method applies a probabilistic ladder network trying to learn conditional Gaussian distributions by forcing different latent features to approximate different Gaussian models. The reconstruction error and the probability of the test sample locating in the latent space are combined to detect the unknown class.
The prototype learning is integrated into the OSR problem in~\cite{yang2020convolutional}, which inspires subsequent prototype-based methods, such as
reciprocal points learning (RPL)~\cite{DBLP:conf/eccv/ChenQ0PLHP020},
adversarial motorial prototype framework (AMPF)~\cite{xia2021adversarial},
and prototype mining and learning(PMAL)~\cite{lu2022pmal}.
The basic idea of these methods is to learn one or more prototypes in an embedded space to represent each known or unknown class. 
The distance to the prototypes can be employed for detecting novel classes.
Recently, 
it has been pointed out that 
the open set performance can be further boosted by improving its closed set accuracy, which demonstrates 
a high correlation between open set and closed set performance~\cite{DBLP:conf/nips/VaswaniSPUJGKP17}.

\subsubsection{Generalized Out-of-Distribution Detection using Vision Transformer}
Open set recognition (OSR) and related problems such as 
anomaly detection (AD), novelty detection (ND),  out-of-distribution detection (OOD), and outlier detection (OD) 
are collectively referred to as generalized out-of-distribution detection~\cite{DBLP:journals/corr/abs-2110-11334}.
Recently, the 
vision transformer (ViT) is used for generalized out-of-distribution detection tasks, for example, 
as a feature extractor for anomaly detection~\cite{DBLP:conf/isie/MishraVFPF21,DBLP:journals/corr/abs-2111-07677}.
Different techniques are used based on the extracted features to identify and localize 
anomalies in industrial images.
The AD task is under the unsupervised setting and no labels are provided in the training data, making it distinctly separate from the OSR task.
OODformer is an OOD detection architecture based on ViT~\cite{koner2021oodformer} which
incorporates the transformer as a feature extractor and performs the detection task by using 
both class-conditioned latent space similarity and a network confidence score.
This approach is closer to our work which employs an additional detection head for improving the detection performance. Further, we focus on different detection tasks, namely, 
OSR has the discriminative power within the known classes
and
specifically targets semantic novelty compared with OOD~\cite{DBLP:journals/corr/abs-2110-11334}.


	\section{Open Set Recognition using Vision Transformer}
We start this section by providing the formal definition of the open set recognition (OSR) problem. 
A brief background on vision transformer (ViT) is then presented, which is the basis of our 
proposed approach.
Following the introduction to preliminaries, we elaborate on our method for OSR using ViT.

\subsection{Problem Definition}
\label{sec: problem}
Consider a training set of labeled samples $\mathcal{D}_\text{train} = \{ (x_i,  y_i)\}_{i=1}^{n} $ from $K$ different known classes,
where each example pair $ (x_i,  y_i)$ contains an input $x_i \in \mathcal{X} \subseteq \mathbb{R}^d$ and corresponding label $y_i\in\mathcal{K} = \{0, \ldots, K-1\}$. 
In an open world setting or an open set scenario,  test data can be drawn not only from the set of known classes $\mathcal{K}$, but also from the set of unknown classes $\mathcal{U} = \{K, \ldots, K+U-1\}$, where $U$ is the number of unknown classes. 
The OSR problem aims to learn a model that performs two sub-tasks: 
(1) novelty detection for unknown classes and (2) classification for known classes.
Specifically, given an unlabeled test sample $x$, the learned model first
identifies whether the test sample is drawn from the set of known classes $\mathcal{K}$.
If $x$ is believed to come from a known class, the model should also output this specific class; otherwise the model just states that $x$ does not belong to any known classes.

\subsection{Vision Transformer}
\label{sec:vit}
Our approach leverages the vision transformer (ViT)~\cite{DBLP:conf/iclr/DosovitskiyB0WZ21} to extract features for classification and detection tasks. 
The ViT model is depicted on the left side of Fig.~\ref{fig:architecture}.
The standard transformer encoder can only receive a $1$D sequence of tokens as input. 
To apply the transformer to a $2$D image, we need to pre-process the image as follows.
The original image $x\in\mathbb{R}^{H\times W\times C}$ is first reshaped into a sequence of 2D patches $x_p\in\mathbb{R}^{N\times (P^2 \cdot C)}$, where $(H, W)$ and $C$ are the resolution and the number of channels of the original image respectively, $(P, P)$ is the resolution of the image patch, and naturally, $N=HW/P^2$ is the number of patches or the size of the sequence. Subsequently, each image patch $x_p^i : i \in \{1, \ldots, N\}$ is mapped to a $1$D $D$-dimensional patch embedding $z_0^i\in\mathbb{R}^D$
through a learnable linear projection $E\in\mathbb{R}^{(P^2\cdot C)\times D}$.
A learnable [class] token $x_\text{cls} \in \mathbb{R}^D$ ($x_\text{cls} = z_0^0$) is prepended to the sequence of patch embeddings, 
and its corresponding output from the last layer of the transformer encoder is utilized for classification.
The positional information is embedded into a sequence of learnable 1D vectors $E_{\text{pos}}\in\mathbb{R}^{(N+1)\times D}$ and added to the patch embeddings to serve as the input to the encoder transformer, which can be formally defined as  
\begin{equation}
\begin{split}
	z_0 = & [z_0^0; z_0^1; \ldots; z_0^N] + E_\text{pos} \\
	    = & [x_\text{cls}; x_p^1 E; \ldots; x_p^N E] + E_\text{pos}.
\end{split}
\end{equation}

The transformer encoder is formed by alternately stacking multiple encoder layers with the same structure, and
the core of each layer is a multiheaded self-attention (MSA) block. 
MSA associates every patch in the input sequence to every other patch, which enables the encoder layer to integrate the information from the global context rather than the local one, and therefore is beneficial to 
the classification.
More details of MSA and the structure of the encoder layer can be found in \cite{DBLP:conf/nips/VaswaniSPUJGKP17} and \cite{DBLP:conf/iclr/DosovitskiyB0WZ21}.
By feeding the embedded input patches $z_0$ through an $L$-layer transformer encoder, 
a sequence of output patches $z_L = [z_L^0; z_L^1; \ldots; z_L^N]$ is obtained from the last layer, where each output patch $z_L^i, i\in\{0, 1, \ldots, N\}$ has the same dimension $D$ as the input patch $z_0^i$.
The state of [class] token $z_L^0$ from the last layer aggregates rich information from the image patch tokens globally, thereby encoding discriminative attributes among objects. 
Such a feature space is denoted as $\mathcal{F} \subseteq \mathbb{R}^D$, 
to which a classification head implemented by a multi-layer perception (MLP) is attached
for the classification task.

\subsection{Proposed Method}
\begin{figure}
	\centering
	\tikzsetnextfilename{architecture}%
	{
		\input{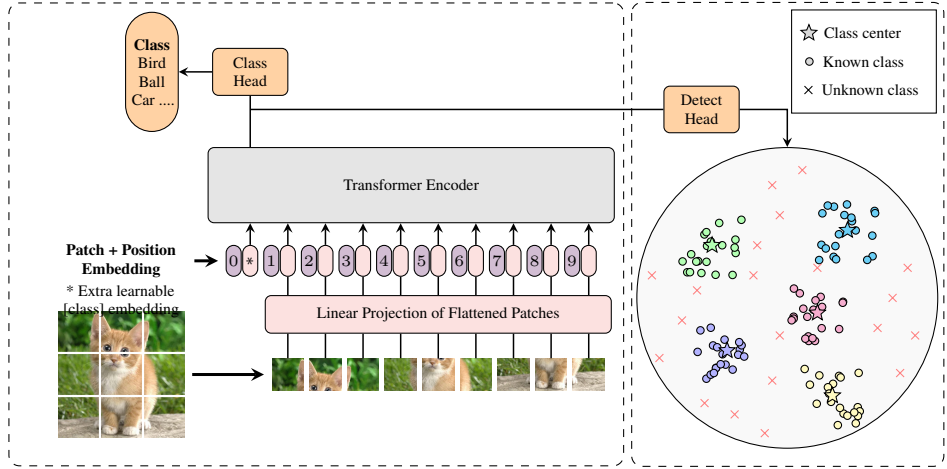}}
	\caption{Proposed architecture for open set recognition. Left: A ViT model to perform classification task for known classes; Right: A detection head attached to ViT to perform novelty detection for unknown classes}
	\label{fig:architecture}
\end{figure}

As described in Sect.~\ref{sec: problem}, the OSR problem consists of two sub-tasks: (1) novelty detection for unknown classes and (2) classification for known classes.
The ViT architecture has already achieved state-of-the-art performance on many image classification benchmarks~\cite{DBLP:conf/iclr/DosovitskiyB0WZ21}.
We keep the original ViT architecture unchanged to retain its advantage of classification task on the closed set. 
In addition, 
an extra detection head is attached to the feature space $\mathcal{F}$
in order to perform the detection task.
The overall architecture of the proposed network is illustrated in Fig.~\ref{fig:architecture}.
The training procedure is divided into two stages for performing the classification and detection tasks respectively.
In the following, we present each training stage and explain how to use the trained model for OSR problem. 

\subsubsection{Closed set training (Stage 1)}
The first training phase is to learn a ViT network to perform the classification task on the closed set.
ViT can be decomposed into two modules:
(1) a feature extraction module  $\phi_f(x; \mathcal{W}_f)$ mapping the input image $x \in \mathcal{X} \subseteq \mathbb{R}^{H\times W\times C}$ to embedded feature $ f \in \mathcal{F} \subseteq \mathbb{R}^D$,
and (2) a classification module $\phi_c(f; \mathcal{W}_c)$ mapping embedded feature  $f$ to class prediction probabilities $p \in \mathcal{P} \subseteq \mathbb{R}^K$, where $\mathcal{W}_f$ and $\mathcal{W}_c$ are the learnable parameters of the modules.

Let  $\mathcal{D}_\text{train} = \{ (x_i,  y_i)\}_{i=1}^{n}$ be a training dataset sampled from $K$ known classes, where $x_i \in \mathcal{X}$ and $y_i \in \mathcal{K} = \{0, \ldots, K-1\}$.
We define the closed set classification objective as
\begin{equation}
	\min_{\mathcal{W}_f, \mathcal{W}_c} -\frac{1}{n} \sum_{i=1}^{n} \sum_{k=0}^{K-1} \mathbb{I}(y_i=k) \log p_i^k,
\end{equation}
where $\mathbb{I}(\text{condition})$ is binary indicator ($0$ or $1$) returning if the [condition] is true, $p_i = \phi_c( \phi_f(x_i; \mathcal{W}_f)  ;\mathcal{W}_c)$ is the prediction probabilities, and $p_i^k$ is the probability of the $i^\text{th}$ image being sampled from the class $k$.

\subsubsection{Open set training (Stage 2)}
The training objective in the second stage is to represent the known class data in a detection space where the representations form tight and class-specific clusters.
The useful embedded features learned from the first training stage can be employed to benefit this objective.
Specifically, 
a detection module $\phi_d(\cdot; \mathcal{W}_d)$ with learnable parameters $\mathcal{W}_d$ is attached to the feature space $\mathcal{F}$, attempting to map the embedded feature $f$ to a representation $e$ in the detection feature space $\mathcal{E} \subseteq \mathbb{R}^D$.   

During the open set training phase, 
the feature extraction parameters $\mathcal{W}^*_f$ and classification module parameters $\mathcal{W}^*_c$  learned from training stage 1 are fixed to maintain the classification accuracy for the known classes. 
Given the training dataset $\mathcal{D}_\text{train} = \{ (x_i,  y_i)\}_{i=1}^{n}$, we optimize the parameters $\mathcal{W}_d$ of the detection module by minimizing the distance between the representation $e_i = \phi_d(\phi_f(x_i; \mathcal{W}^*_f); \mathcal{W}_d)$ and its class center $c_{y_i}$, 
which can be formally described as 
\begin{equation}
	\min_{\mathcal{W}_d} \frac{1}{n} \sum_{i=1}^{n} = \Vert e_i - c_{y_i}   \Vert^2 .
\end{equation}

Inspired by the center strategy in~\cite{DBLP:conf/icml/RuffGDSVBMK18}, for each known class $k \in \mathcal{K}$, we anchor its center as the mean of the representations that are computed from the initial forward pass on the training data from class $k$.
After the training, the data from known classes are encouraged to be closely mapped to class-specific centers in the detection space, 
whereas the data from unknown classes lie far away from the centers.
It should note that the class label $y_i$ is used to specify the class center $c_{y_i}$, and therefore, the optimization procedure of open set training is also supervised.

\subsubsection{Testing}
After the network training procedure, we obtain the network modules $\phi_f$, $\phi_c$, $\phi_d$ and their learned parameters $\mathcal{W}_f^*$, $\mathcal{W}_c^*$, $\mathcal{W}_d^*$. 
At test time, the trained network should not only correctly classify the known classes but also reject the unknown classes.

For a new test example $x$, its embedded feature $f$ is extracted by feeding it into the feature extraction module, $f = \phi_f(x; \mathcal{W}_f^*)$.
Its predicted class label $\hat{y}$ corresponds to the highest probability score in the predicted probabilities 
\begin{equation}
\hat{y} = \argmax(p) = \argmax(\phi_c(f; \mathcal{W}_c^*)).	
\end{equation}
The embedded feature $f$ of input $x$ is also mapped to the detection space as $e$, and the Euclidean distance to its corresponding cluster center $c_{\hat{y}}$ is naturally defined as  the anomaly score
\begin{equation}
s = \Vert e - c_{\hat{y}} \Vert ^ 2.
\end{equation}
If the anomaly score is greater than a predefined threshold $\tau$, the input $x$ is rejected as an unknown example;
otherwise, the model predicts $x$ as known with label $\hat{y}$.
The detailed testing procedure is summarized in Algorithm~\ref{alg:detection_algorithm}.

\begin{algorithm}
	\caption{Open set recognition using vision transformer. }
	\label{alg:detection_algorithm}
	\begin{algorithmic}[1]
		\Require Test example $x$
		\Require Network modules $\phi_f$, $\phi_c$, $\phi_d$ and their trained parameters $\mathcal{W}^*_f$, $\mathcal{W}^*_c$, $\mathcal{W}^*_d$
		\Require A set of cluster centers in detection space for $K$ known classes $\{c_0, \ldots, c_{K-1}\}$   
		\Require Detection threshold $\tau$ 
		\State Embedded feature $f = \phi_f(x; \mathcal{W}_f^*)$
		\State Predicted known label $\hat{y} = \argmax(\phi_c(f; \mathcal{W}_c^*))$
		\State Representation in detection space $e = \phi_d(f; \mathcal{W}_d^*) $
		\State Anomaly score $s = \Vert e - c_{\hat{y}} \Vert ^ 2$
		\If{$s > \tau$}
		\State Predict $x$ as unknown
		\Else
		\State Predict $x$ as known with label $\hat{y}$		
		\EndIf
	\end{algorithmic} 
\end{algorithm}
	\section{Evaluation}
\tikzexternalize[prefix=tikz/]
This section presents a comprehensive evaluation of the proposed approach and comparison with state-of-the-art methods.
The experiments are not limited to the standard benchmarks in the OSR literature, 
but additional datasets such as large-scale, long-tailed, and fine-grained datasets are adopted to demonstrate the broad effectiveness of our approach.    
All experiments reported in this paper were conducted on an $80$-core Ubuntu Linux virtual machine with 
$\SI{128}{\giga\byte}$ RAM and four Tesla V100 GPUs.

\subsection{Implementation Details}
The proposed architecture is based on the ViT, and we use the ViT-B/$16$ variant as the backbone network. 
ViT model is typically pre-trained with a large dataset and then fine-tuned on downstream tasks.
Therefore, as suggested in~\cite{DBLP:conf/iclr/DosovitskiyB0WZ21},
all models used in our experiments were initially pre-trained on the ImageNet-21K~\cite{DBLP:journals/ijcv/RussakovskyDSKS15} dataset.  
During the fine-tuning closed set training phase, the original classification head is removed, and a single linear layer with a SoftMax activation function is attached,
whose output dimension corresponds to the number of known classes.
After closed set training, the parameters in ViT are fixed. An additional detection head with a single linear layer is attached, where the dimension of detection feature space is the same as the feature dimension $D=768$ in ViT.   
We employ an SGD optimizer with a learning rate of $0.01$ and momentum of $0.9$ for both closed set and open set training.
Each training phase takes $4590$ steps with batch size $256$.

\subsection{Evaluation Metrics}
Following the standard evaluation protocols in the OSR literature, 
the closed set and open set performance are evaluated separately.
For closed set evaluation, we report top-$1$ classification accuracy on test data for known classes.
The receiver operating characteristic curve
plots the true positive rate against the false positive rate by varying the detection threshold.
The area under the receiver operating characteristic (AUROC) curve is a threshold-independent metric that can be interpreted as the probability that an
example from a known class is assigned a lower anomaly score than an example from an unknown class~\cite{DBLP:journals/prl/Fawcett06}.
Therefore, the AUROC is adopted for open set performance evaluation.

\subsection{Experiments on Standard Benchmarks}

\subsubsection{Datasets}
In order to fairly compare with the start-of-the-art OSR methods, we evaluate our approach using standard benchmark datasets and protocols, which are briefly introduced below. 
\begin{itemize}
	\item\textbf{MNIST, SVHN, CIFAR10.} MNIST~\cite{lecun1998gradient}, SVHN~\cite{netzer2011reading}, and CIFAR10\cite{krizhevsky2009learning} are 10-class datasets. 
	$6$ classes are randomly sampled  as known classes, while the remaining $4$ classes are regarded as unknown classes.
	\item \textbf{CIFAR+\textit{N}.} For CIFAR+$N$ experiments, $4$ classes are randomly selected from CIFAR10 as known classes, and $N$ non-overlapping classes from CIFAR100~\cite{krizhevsky2009learning} are used as unknown classes, where $N$ can be either $10$ or $50$.
	\item \textbf{TinyImageNet.} TinyImageNet dataset is a subset of the ImageNet~\cite{DBLP:journals/ijcv/RussakovskyDSKS15} dataset and totally contains $200$ classes. $20$ classes are randomly sampled as known classes, while the remaining $180$ classes are considered as unknown classes.
\end{itemize}

\subsubsection{Comparison with State-of-the-Art}
We compare our proposed approach with other OSR methods in the literature, including SoftMax~\cite{DBLP:conf/eccv/NealOFWL18}, C2AE~\cite{DBLP:conf/cvpr/OzaP19}, OpenHybrid \cite{DBLP:conf/eccv/0002LG020}, ARPL~\cite{chen2021adversarial}, AMPF++~\cite{xia2021adversarial}, and PMAL~\cite{lu2022pmal}.
As described before, we measure the classification accuracy and AUROC to evaluate the closed set and open set performance.
The comparison results of both closed set and open set performance  are presented in Table~\ref{tab:standard_dataset}.
The reported results are averaged among five random ``known/unknown" splits.

As illustrated in the table,  
the ViT-based method obtains the best closed set classification accuracy on all benchmarks except the MNIST dataset. Especially on TinyImageNet, the accuracy improvement reaches $11.2\%$.

As for the open set performance, 
new state-of-the-art results are obtained for the CIFAR10, CIFAR+10, CIFAR+50, and TinyImageNet 
datasets, and 
the AUROC improvements are $4.1\%$, $1.6\%$, $2.4\%$, and $14.6\%$, respectively. 
An interesting result shown in the table is that the AUROC of our approach on MNIST and SVHN datasets are not comparable with the other methods, which may be due to the small size of the MNIST and SVHN datasets.
The detection performance is consistent with the classification performance of the ViT, which does not seem to have an advantage on small-scale training datasets.

\begin{table}
\caption{Closed set accuracy and open set AUROC for comparison of our approach with 
state-of-the-art methods on the open set recognition standard benchmark datasets.
All values are percentages and the best results are highlighted in \textbf{bold}.
The results other than our approach are obtained from \cite{lu2022pmal} and \cite{xia2021adversarial}.
We denote `C' for `CIFAR' and `Tiny' for `TinyImageNet'
}
\label{tab:standard_dataset}
\begin{center}
	\scriptsize
	\begin{tabularx}{\textwidth}{c *{12}{Y}}
		\toprule
		\multirow{3}*{Methods}
		& \multicolumn{6}{c}{Closed Set Accuracy}  
		& \multicolumn{6}{c}{Open Set AUROC}\\
		\cmidrule(r){2-7} \cmidrule{8-13}
		& \tiny MNIST & \tiny SVHN & C10 & C+10 & C+50 & Tiny 
		& \tiny MNIST & \tiny SVHN & C10 & C+10 & C+50 & Tiny\\
		\midrule
		SoftMax  & 99.5 &  94.7 &  80.1 & 96.3 & 96.4 & 72.9 & 97.8 &  88.6 &  67.7 & 81.6 & 80.5 & 57.7\\
		C2AE  & - & - & - & - & - & - & 98.9 & 92.2 & 89.5 & 95.5 & 93.7 & 74.8 \\
		OpenHybrid  & 94.7 & 92.9 & 86.8 & - & - & - & 99.5 & 94.7 & 95.0 & 96.2 & 95.5 & 79.3 \\
		ARPL  & 99.5 &  94.3 &  87.9 & 94.7 & 92.9 & 65.9 & \textbf{99.7} &  96.7 &  91.0 & 97.1 & 95.1 & 78.2\\
		AMPF++  & \textbf{99.8} &  96.9 &  96.0 & 97.5 & 97.4 & 81.1 & \textbf{99.7} &  96.8 &  91.6 & 97.3 & 95.4 & 79.7\\	
		PMAL  & \textbf{99.8} &  97.1 &  97.5 & 97.8 & 98.1 & 84.7 & \textbf{99.7} &  \textbf{97.0} &  95.1 & 97.8 & 96.9 & 83.1\\
		\midrule
		Ours & 99.7 & \textbf{97.3} & \textbf{99.1} & \textbf{99.6} & \textbf{99.6} & \textbf{95.9} & 95.8 & 93.6 & \textbf{99.2} & \textbf{99.4} & \textbf{99.3} & \textbf{97.7}\\
		\bottomrule

	\end{tabularx}
	
\end{center}
	
\end{table}

\subsection{Experiments on Additional Benchmarks}

\subsubsection{Datasets}
Recently, additional special benchmarks have been proposed in the literature for OSR evaluation. 
\begin{itemize}
	\item \textbf{Large-scale dataset: ImageNet-100, ImageNet-200}.  We first demonstrate our approach on a larger scale dataset -- ImageNet-2012~\cite{DBLP:journals/ijcv/RussakovskyDSKS15}, which contains $1$K-classes with more than $1, 000, 000$ images. 
	Following the settings in~\cite{yang2020convolutional}, two experiments, ImageNet-100 and ImageNet-200, separately employ the first $100$ and $200$ classes in ImageNet as knowns, and the rest classes are regarded as unknowns. 
	\item \textbf{Fine-grained dataset: CUB}. 
	In this experiment, we consider a more complicated scenario -- fine-grained visual categorization (FGVC) dataset, in which all classes are variants of a single category rather than distinct categories compared to existing OSR benchmarks. 
	The fine-grained discrimination among classes makes both classification and detection tasks more difficult.
	Caltech-UCSD Birds (CUB) 200 dataset~\cite{WelinderEtal2010} 
	is a typical FGVC dataset containing the photos of 200 bird species. It is utilized to evaluate OSR methods in~\cite{vaze2021open}.      
	We choose $100$ bird species as knowns, and the remaining $100$ species serve as unknowns.  
	The $100$ unknown species are further subdivided into three classes ``Easy", ``Medium", and ``Hard" according to the detection difficulty, consisting of $32$, $34$, and $34$ classes respectively.
	More details about this experimental protocol can be found in~\cite{vaze2021open}.
	\item \textbf{Long-tailed dataset: ImageNet-LT}. 
	In the real world, the frequency distribution of the classes is normally not balanced but long-tailed, with a few common and many rare categories~\cite{DBLP:conf/cvpr/0002MZWGY19}. 
	ImageNet-LT dataset is a long-tailed dataset constructed by unevenly sampling from the categories of the original ImageNet-2012 dataset. In summary, it totally has 115.8K images from 1000 categories,
	with maximally 1280 images per class and minimally 5 images per class. 
	The whole training dataset from ImageNet-LT is used as knowns,  
	while additional classes from ImageNet-2010 are utilized in the testing phase as the unknown classes.  
	
\end{itemize} 

\subsubsection{Evaluation Results}
Table~\ref{tab:additional_dataset} reports the results of our proposed approach compared with the other OSR methods on these three additional datasets. 
The methods used for comparison include
SoftMax~\cite{DBLP:conf/eccv/NealOFWL18},CPN~\cite{yang2020convolutional}, RPL~\cite{DBLP:conf/eccv/ChenQ0PLHP020}, MPF~\cite{xia2021adversarial}, PMAL~\cite{lu2022pmal}, SoftMax+~\cite{vaze2021open}, and ARPL+~\cite{vaze2021open}.
Since the ``known/unknown" splits in these datasets are fixed, the reported results are obtained from a single experimental trial.

As can be seen from the table, the proposed approach significantly improves the closed set classification accuracy for all three datasets.
We should note that the improvement on the ImageNet-LT dataset is $41.8\%$. This is largely because the ViT is pre-trained on ImageNet21K, allowing a high accuracy even on some rare categories. 
The approach outperforms the other methods and achieved new state-of-the-art results on the novelty detection task.
The AUROCs have increased by $4.9\%$, $5.5\%$, and $22.0\%$ on ImageNet-100, ImageNet-200, and ImageNet-LT datasets, respectively. The minimum improvement in the three subsets of the CUB dataset has also reached  $7.2\%$.

\begin{table}
	
	\caption{Closed set accuracy and open set AUROC for comparison of our approach with state-of-the-art methods on the additional benchmark datasets.
	The values other than our approach are obtained from~\cite{lu2022pmal},\cite{vaze2021open}, and \cite{xia2021adversarial}. 
		We denote `IN' for `ImageNet'
	}
	\label{tab:additional_dataset}
	\begin{center}
		\scriptsize
		\begin{tabularx}{\textwidth}{c *{9}{Y}}
			\toprule
			\multirow{3}*{Methods}
			& \multicolumn{4}{c}{Closed Set Accuracy}  
			& \multicolumn{5}{c}{Open Set AUROC}\\
			\cmidrule(r){2-5} \cmidrule{6-10}
			& IN-100 & IN-200  & IN-LT & CUB
			& IN-100 & IN-200 & IN-LT & \multicolumn{2}{c}{CUB} \\
			\midrule
			SoftMax  & 81.7 &  79.7 & 37.8 & - & 79.7 & 78.4 &  53.3  & \multicolumn{2}{c}{-} \\
			CPN  &  86.1 & 82.1 & 37.1 & - & 82.3 &  79.5 &  54.5 & \multicolumn{2}{c}{-} \\
			RPL  & 81.8 &  80.7 & 39.0 & - & 81.2 & 80.2 &  55.1  & \multicolumn{2}{c}{-} \\
			MPF  & 80.8 &  83.8 & - & - & 94.6 & 95.6 &  -  & \multicolumn{2}{c}{-} \\		
			PMAL  & 86.2 &  84.1 & 42.9 & - & 94.9 & 93.9 &  71.7  & \multicolumn{2}{c}{-} \\
			SoftMax+  & - &  - & - & 86.2 & - &  - &  - & \multicolumn{2}{c}{88.3/82.3/76.3} \\
			ARPL+  & - &  - &  - & 85.9 & - &  - &  - & \multicolumn{2}{c}{83.5/78.9/72.1} \\
			\midrule
			Ours & \textbf{87.3} &  \textbf{87.5} &  \textbf{84.7} & \textbf{93.9} & \textbf{99.8} &  \textbf{99.4} &  \textbf{93.7} & \multicolumn{2}{c}{\textbf{93.3/87.3/79.3}} \\
			\bottomrule
			
		\end{tabularx}
		
	\end{center}
	
\end{table}

\subsection{Further Discussion}
\subsubsection{Visualization on Embedded Space}
\begin{figure}
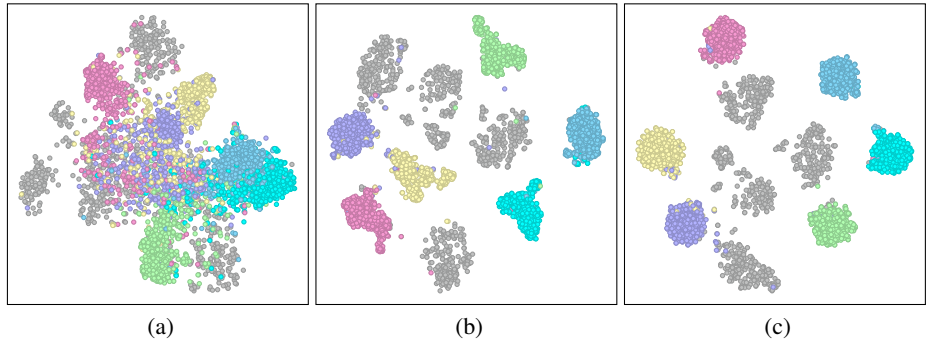

	\centering
		\tikzsetnextfilename{tsne1}%
		\subfloat[]{\label{fig:tsne_1}\input{./figures/tsne1.tex}}
		\tikzsetnextfilename{tsne2}%
		\subfloat[]{\label{fig:tsne_2}\input{./figures/tsne2.tex}}
		\tikzsetnextfilename{tsne3}%
		\subfloat[]{\label{fig:tsne_3}\input{./figures/tsne3.tex}}
	\label{fig:tsne}
	\caption{The learned embedded features visualized by 2D T-SNE. (a) embedded space of ViT after pre-training; 
		(b) embedded space of ViT after closed set training; (c) detection feature space of proposed approach after open set training.
		Colored dots in the figures represent the knowns while gray dots represent unknowns. The results are obtained from a trial on CIFAR10, where $6$ known classes are \{airplane, automobile,  cat, deer, dog,  truck\}, and $4$ unknown classes are \{bird, frog, horse, ship\} }
\end{figure}

In order to gain a deeper understanding of the proposed approach, we utilize our trained model to map $4000$ test data from known classes and $800$ test data from unknown classes to the detection space, 
and plot the 2D T-SNE latent representations in this space in Fig.~\ref{fig:tsne_3}.
We observe that the data from $6$ known classes are mapped to $6$ tight clusters in the detection space, and there is little overlap between clusters. The representations of unknown classes are clearly separated from $6$ known clusters, which is very helpful for novelty detection of unknown classes in the detection space.  
Although the training set does not contain any data from $4$ unknown classes, we can vaguely see that the data from $4$ unknown classes form $4$ loose clusters in the detection space. 
This may be because ViT is pre-trained on the large dataset -- ImageNet-21K, giving the model a certain ability to segment objects. 

\subsubsection{Ablation Study}
We perform an ablation analysis on a CIFAR10 trial to evaluate the contribution of each individual module and training phase to the overall performance. 
To illustrate the effect of pre-training on our approach,
we first evaluate a ViT model that was only pre-trained on ImageNet-21K but not trained on the training dataset. 
The 2D latent representations in the feature space of ViT are shown in Fig.~\ref{fig:tsne_1}. 
As we can see,  the distribution of known and unknown data representations is more random and there is a great overlap between classes. Yet, the data from the same class are still clustered. 
Even though the model is not trained on the training dataset, the open set AUROC of this ViT model is $87.40\%$, which surpasses the SoftMax method and is very close to C2AE method.
It should be noted that we use the distance to its closest cluster center in the feature space as the anomaly score when directly using this ViT model to measure the AUROC. 

Then, we evaluate the ViT model after training stage 1 -- open set training. We plot the 2D feature space of ViT in Fig.~\ref{fig:tsne_2}. Compared with Fig.~\ref{fig:tsne_1}, the known classes are separated from the unknown classes with little overlap in the feature space. 
Taking the distance to the class-specific cluster center  as the anomaly score, the open set AUROC reaches $99.0\%$, which outperforms all the other methods in the literature. 

Finally, 
we evaluate the proposed approach by attaching an additional detection head to the previous ViT classification model and perform the training stage 2 -- closed set training. After the training, the representations in the detection space are shown in Fig.~\ref{fig:tsne_3}. Compared to Fig.~\ref{fig:tsne_2}, the clusters of known classes are more compact. The AUROC of our proposed approach on this trial exceeds the result after training stage 1 and reaches $99.5\%$. This is because, for the known classes, the distances to the cluster centers become smaller 
through the training stage 2. 
This experiment illustrates the detection head and training phase 2 can further improve the detection performance and help proposed approach achieve state-of-the-art results.

	\section{Conclusions}
This paper presents an approach to exploit the vision transformer (ViT) for performing open set recognition (OSR). 
We use the ViT network to perform the closed set classification problem 
and incorporate an additional detection head to detect novel classes. 
The comprehensive evaluation shows  
that our approach surpasses other methods and improves the state-of-the-art results on almost all the OSR benchmarks by a considerable margin.
The proposed method relies heavily on supervised learning to extract the discriminative features for both classification and detection tasks. It hinders the method from applying to other generalized out-of-distribution detection problems, e.g., anomaly detection, which does not provide the class labels.
Addressing this limitation, for example, using self-supervised learning to extract features, is an interesting research topic for future work.
	
	\bibliographystyle{plain}
	\bibliography{egbib}
	
\end{document}